\documentclass[10pt,twocolumn,letterpaper]{article}

\usepackage{cvpr}              % To produce the CAMERA-READY version
\usepackage{graphicx}
\usepackage{amsmath}
\usepackage{amssymb}
\usepackage{booktabs}

\usepackage{multirow}
\usepackage{algorithm}
\usepackage{algpseudocode}
\usepackage{amsmath}
\usepackage{graphics}
\usepackage{epsfig}
\usepackage[title]{appendix}

\usepackage[pagebackref,breaklinks,colorlinks]{hyperref}

\usepackage[capitalize]{cleveref}
\crefname{section}{Sec.}{Secs.}
\Crefname{section}{Section}{Sections}
\Crefname{table}{Table}{Tables}
\crefname{table}{Tab.}{Tabs.}

\def\cvprPaperID{3434} % *** Enter the CVPR Paper ID here
\def\confName{CVPR}
\def\confYear{2022}

\begin{document}

\title{Modality-Aware Triplet Hard Mining for Zero-shot \\Sketch-Based Image Retrieval }

\author{Zongheng Huang\textsuperscript{1}, Yifan Sun\textsuperscript{2}, Chuchu Han\textsuperscript{1}, Changxin Gao\textsuperscript{1}, Nong Sang\textsuperscript{1}\\
% Huazhong University of Science and Technology\\
% \and
% YiFan Sun\\
% %First line of institution2 address\\
% \and
% Chuchu Han\\
% % Huazhong University of Science and Technology\\
% %Institution1 address\\
% \and
% Changxin Gao\\
% % Huazhong University of Science and Technology\\
% %Institution1 address\\
% \and
% Nong Sang\\
\textsuperscript{1}Key Laboratory of Ministry of Education for Image Processing and Intelligent Control, \\School of Artificial Intelligence and Automation, Huazhong University of Science and Technology,\\
\textsuperscript{2}Baidu Research\\
{\tt\small \{huangzongheng, hcc, cgao, nsang\}@hust.edu.cn, }
{\tt\small sunyf15@tsinghua.org.cn}
% {\tt\small huangzongheng@hust.edu.cn}
% {\tt\small sunyf15@tsinghua.org.cn}
% {\tt\small hcc@hust.edu.cn}
% {\tt\small  cgao@hust.edu.cn}
% {\tt\small  nsang@hust.edu.cn}
}

\maketitle

%%%%%%%%% ABSTRACT
\begin{abstract}
   This paper tackles the Zero-Shot Sketch-Based Image Retrieval (ZS-SBIR) problem from the viewpoint of cross-modality metric learning. % with recent good practices in deep metric learning. 
   This task has two characteristics: 1) the zero-shot setting requires a metric space with good within-class compactness and the between-class discrepancy for recognizing the novel classes and 2) the sketch query and the photo gallery are in different modalities. 
   The metric learning viewpoint benefits ZS-SBIR from two aspects. 
   First, it facilitates improvement through recent good practices in deep metric learning (DML). By combining two fundamental learning approaches in DML, \emph{e.g.}, classification training and pairwise training, we set up a strong baseline for ZS-SBIR. Without bells and whistles, this baseline achieves competitive retrieval accuracy. Second, it provides an insight that properly suppressing the modality gap is critical. To this end, we design a novel method named Modality-Aware Triplet Hard Mining (MATHM). MATHM enhances the baseline with three types of pairwise learning, \emph{e.g.}, a cross-modality sample pair, a within-modality sample pair, and their combination.
   We also design an adaptive weighting method to balance these three components during training dynamically. 
   Experimental results confirm that MATHM brings another round of significant improvement based on the strong baseline and sets up new state-of-the-art performance. For example, on the TU-Berlin dataset, we achieve 47.88+2.94\% mAP@all and 58.28+2.34\% Prec@100.
   Code will be publicly available.
	\vspace{-0.5cm} 
\end{abstract}
 
%%%%%%%%% BODY TEXT
\section{Introduction}

Sketch-Based Image Retrieval (SBIR) aims to retrieve the image-of-interest in the database, using a hand-drawn sketch as the query. It allows searching without image queries and brings great convenience to potential users. The conventional SBIR requires that the training and testing set share the same categories, which limits its real-world applications. In this paper, we consider a more realistic and challenging setting named Zero-Shot SBIR (ZS-SBIR), where all categories of the testing set have not appeared in the training set.

\begin{figure}[!t]
	\centering
	
	\includegraphics[width={0.96\linewidth}]{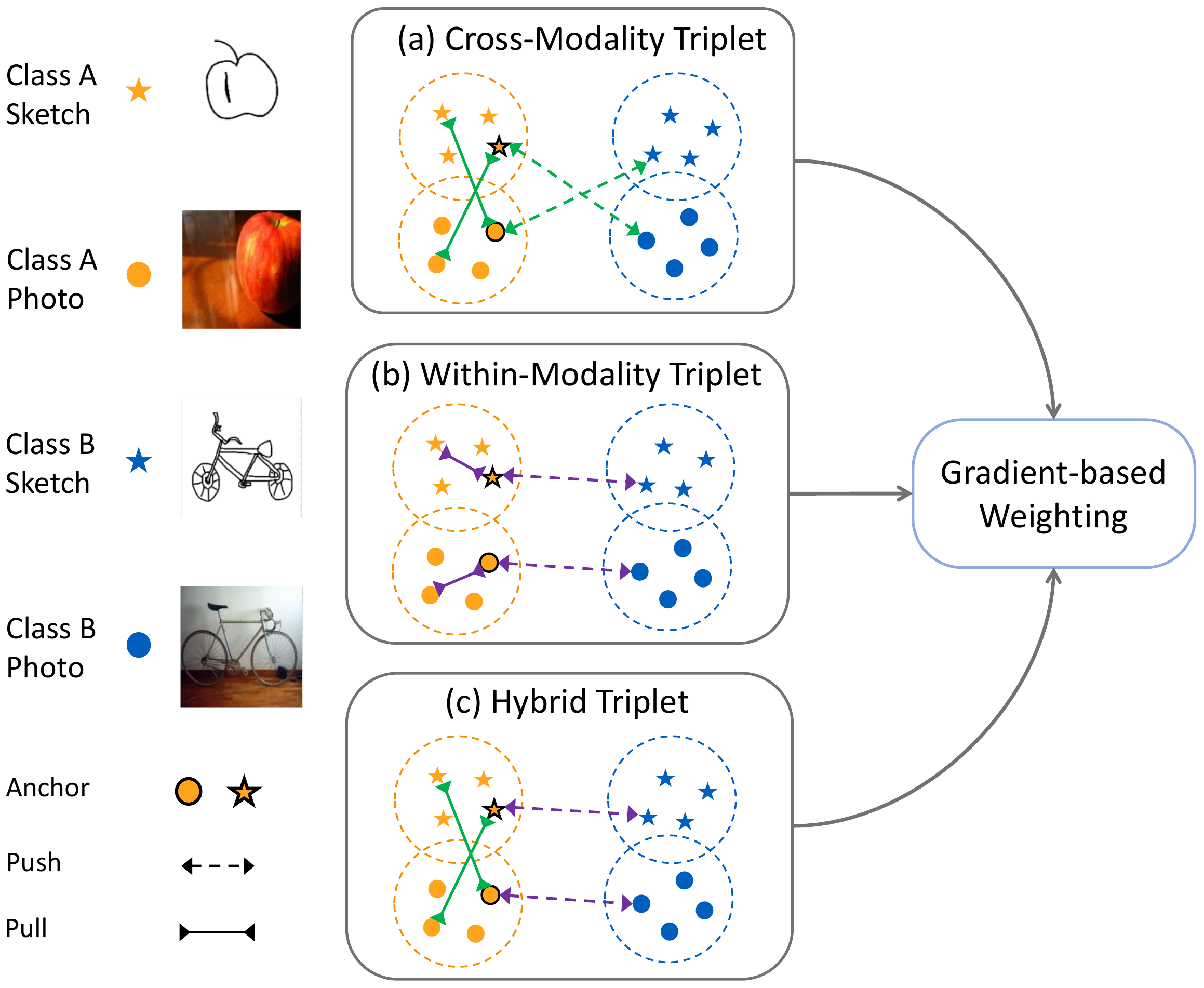}
	\vspace{-0.0cm} 
	\caption{An illustration of three types of triplets. % used for computing triplet losses.   
		(a) positive and negative images have modalities modality from the anchor image;
		(b) positive and negative images have the same modality as the anchor image;
		(c) positive images have different modalities from the anchor image, and negative images have the same modality as the anchor image.
		In contrary to the baseline using only (a) triplets, the proposed MATHM dynamically integrates all the three types of triplets through a novel gradient-based weighting strategy.
	}
	\vspace{-0.1cm} \label{Fig:overview}
\end{figure}

%cross-modality metric learning   within-class compactness and between-class discrepancy
ZS-SBIR can be viewed as a specified cross-modality metric learning task. In such a task, we consider there are two challenges. \emph{1) Learning a metric space with category discrimination.} Specifically, the zero-shot setting is popular in a series of metric learning applications, \emph{e.g.}, fine-grained image retrieval~\cite{6910029, reed2016learning, zhang2016embedding}, face recognition~\cite{taigman2014deepface, schroff2015facenet, wang2017normface, liu2017sphereface, wang2018additive, wang2018cosface, deng2019arcface}, person re-identification~\cite{sun2017svdnet, hermans2017defense, fan2019spherereid} and vehicle re-identification~\cite{chu2019vehicle, li2019vehicle}. In all these tasks, there is no class intersection between the training and testing set, which is consistent with the zero-shot setting. The keynote of these tasks is to learn a metric space in which all the instances from the same class are close and different classes are far away from each other. Such within-class compactness and the between-class discrepancy are vital for the discriminative ability of the novel classes. \emph{2) Suppressing the modality gap between the sketch query and the photo gallery.} Specifically, the query sketches and gallery images belong to different modalities. The hand-drawn sketch usually contains only the object contour that indicates the primary structure. The details are either neglected or distorted. In contrast, real photos include rich texture and color information. % This significant modality gap compromises the within-class compactness and challenges learning an accurate metric space.
This significant modality gap compromises the within-class compactness, which brings the difficulty of building an accurate metric space.

% The metric L_earning viewpoint brings two-fold inspiration for tackling ZS-SBIR, which are explained as follows.
In response to these two challenges, we tackle the ZS-SBIR problem with metric learning techniques from the following two aspects:

% $\bullet$ \emph{First, it benefits ZS-SBIR with recent good practices in deep metric learning (DML).} 
$\bullet$ \emph{First, we set up a strong baseline for ZS-SBIR with recent good practices in deep metric learning (DML) to enhance the category discrimination of feature embedding.} 
There are two fundamental deep metric learning approaches~\cite{sun2020circle, luo2019bag, deng2019arcface}, \emph{e.g.}, classification training and pairwise training. Specifically, the former learns a deep classifier on the training set, encouraging the model to produce class discriminative deep embedding for feature representation. %and uses the deep embedding for feature representation, 
We note that prior works~\cite{lu2018learning, dey2019doodle, yelamarthi2018zero, zhang2020zero, liu2019semantic, zhu2020ocean, deng2020progressive} in ZS-SBIR mainly adopt this learning manner. In contrast, the latter directly compares the features against each other in a pairwise manner. A popular implementation is to construct the feature triplet consisting of an anchor, a positive, and a negative feature. The anchor tries to pull the positive feature close and push the negative feature far away. By adding an additional margin to the feature triplets \cite{hermans2017defense}, the between-class discrepancy of the learned deep embedding is further enlarged.
% the category discrimination of feature representation is further enhanced.

Combining these two fundamental training manners with slight task-specific modification, we set up a strong baseline for ZS-SBIR. 
In the baseline, we naturally modify the ``Pos-Anchor-Neg'' triplet into ``photo-sketch-photo'' form (``$\text{P}^\text{p}\text{-}\text{A}^\text{s}\text{-}\text{N}^\text{p}$'' for simplicity) to meet the objective of the ZS-SBIR task, as shown in Figure~\ref{Fig:overview} (a).
%The modification is that, we construct the ``positive-anchor-negative'' triplet into ``photo-sketch-photo'' form, \emph{e.g.}, both the positive and negative samples are photos and the anchor is the sketch, as illustrated in Fig~\ref{Fig:overview} (a). Such triplet construction is consistent with the objective of ZS-SBIR: given a sketch query, ZS-SBIR ranks the positive photo in front of the negative photo. 
Without any bells and whistles, it achieves retrieval accuracy on par with the state-of-the-art. % While we note that its implementation has limited novelty, 
We believe the simplicity and competitive performance will allow it to serve as a strong baseline for the ZS-SBIR community.	% actually 
%The objective of ZS-SBIR is to retrieve images of the same category as the query sketch, it is crucial to ensure that the distance of the positive pairs ``$\text{P}^\text{p}\text{-}\text{A}^\text{s}$'' are smaller than that of the negative pairs ``$\text{A}^\text{s}\text{-}\text{N}^\text{p}$''. 
%The problem of the above baseline lies in that the ``$\text{P}^\text{p}\text{-}\text{A}^\text{s}\text{-}\text{N}^\text{p}$'' triplet form does not necessarily suppress the modality gap (to be explained in Section \ref{sec:MATHM}). If we provide better suppression on the modality gap, the performance of this baseline can be further improved.

% $\bullet$ \emph{Second, the metric learning viewpoint provides an insight that suppressing the modality gap is crucial. }
% $\bullet$ \emph{Second, we propose a novel triple sampling method named Modality-Aware Triplet Hard Mining (MATHM) to improve triplet ranking loss by suppress the modality gap as well as learning discriminative feature. }

$\bullet$ \emph{Second, we propose a novel method named Modality-Aware Triplet Hard Mining (MATHM) to suppress the modality gap between the sketch query and the photo gallery.} 
%Figure~\ref{Fig:overview} gives an overview of the proposed MATHM method.
%, which suppresses the modality gap without hurting the between class discrepancy within every single modality.
% to meet ZS-SBIR's objective of matching photos to sketch query, as shown in Fig.\ref{Fig:overview} (a). 
%However, we find that this triplet form does not necessarily suppress the modality gap (to be explained in Section \ref{sec:MATHM}). 
%In response to the modality gap problem that is not well address in our baseline (to be explained in Section \ref{sec:MATHM}), 
Based on the baseline, MATHM further narrows the modality gap by supplementing the baseline with two other triplet forms, \emph{e.g.}, the ``$\text{P}^\text{s}\text{-}\text{A}^\text{s}\text{-}\text{N}^\text{s}$'' and ``$\text{P}^\text{p}\text{-}\text{A}^\text{s}\text{-}\text{N}^\text{s}$'' triplets, as illustrated in Figure \ref{Fig:overview} (b) and (c), respectively. The ``$\text{P}^\text{s}\text{-}\text{A}^\text{s}\text{-}\text{N}^\text{s}$'' enforces category discrimination under a single modality, while the ``$\text{P}^\text{p}\text{-}\text{A}^\text{s}\text{-}\text{N}^\text{s}$'' triplet explicitly suppresses the modality gap. Moreover, we design an adaptive weighting scheme to balance these triplets during training dynamically. In Section \ref{Sec:adv-exp}, we reveal an important superiority of the proposed MATHM, \emph{i.e.}, it suppresses the modality gap without decreasing the between-class discrepancy within every single modality.

The contributions are summarized as follows:
\begin{itemize}
%\vspace{-0.2cm}
	% (1) We interpret ZS-SBIR as a cross-modality metric learning problem and construct a simple yet strong baseline by combining two fundamental deep metric learning methods.
	\item We construct a simple yet strong baseline for the ZS-SBIR task by combining two fundamental deep metric learning methods. Without bells and whistles, the performance of our baseline is on par with recent works.
	\vspace{-4mm}
	\item We propose a novel method named Modality-Aware Triplet Hard Mining, which further narrows the modality gap with three types of pair-wise learning. Meanwhile, an adaptive weighting strategy is developed to balance the three components.
% 	\vspace{1mm}
	\item We evaluate our proposed MATHM on two SBIR datasets. It surpasses the previously best methods and sets new state-of-the-art.
% 	\vspace{-1mm}
\end{itemize}

\begin{figure*}[t]
	\centering
% 	\vspace{-0.1cm}
	%\includegraphics[width=\linewidth]{Figures/fig2.pdf}
	\includegraphics[width=0.80\linewidth]{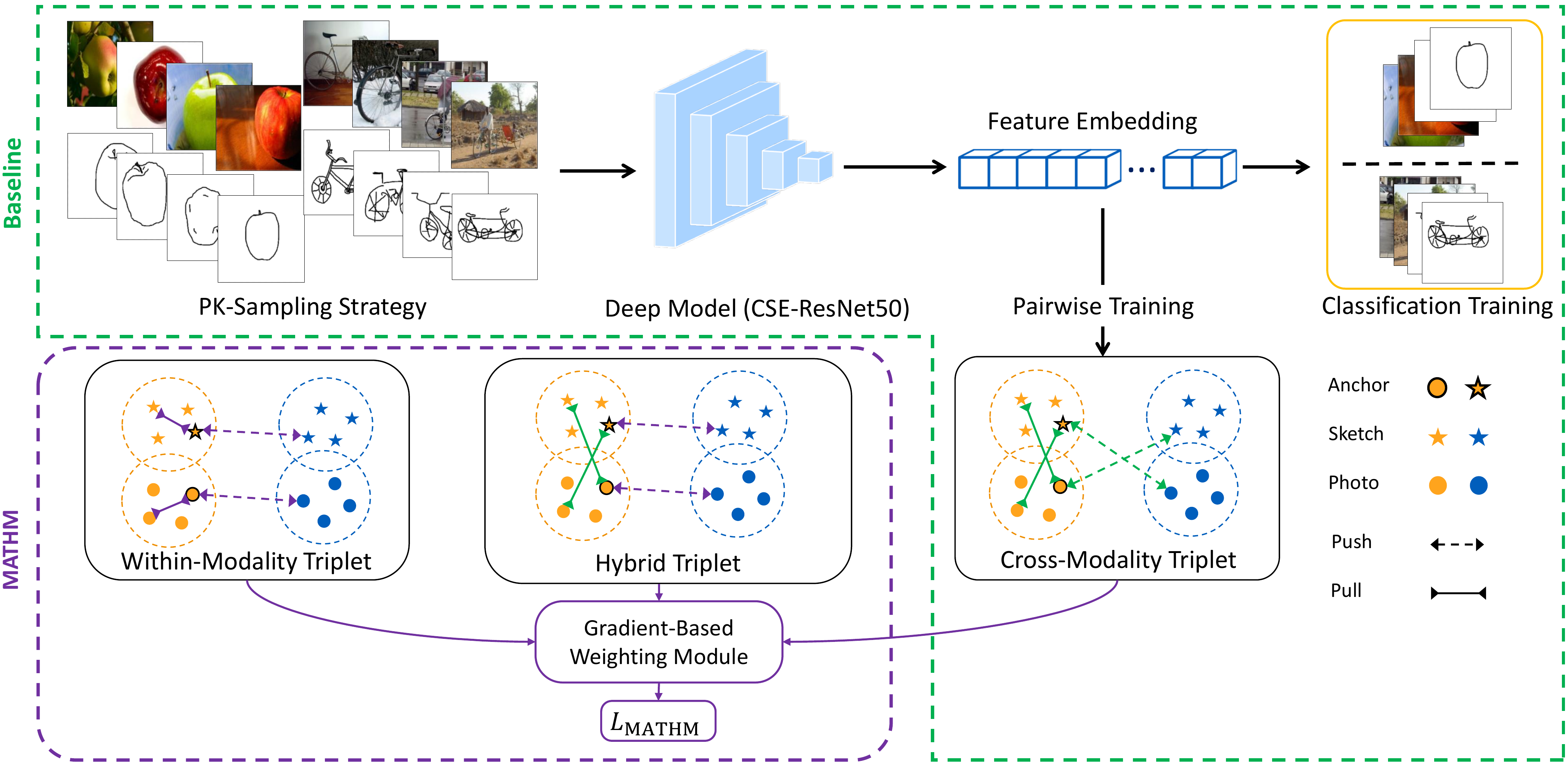}
	\vspace{-0.0cm} \caption{
		An overview of the proposed ZS-SBIR baseline, as well as the novel MATHM method. The \textbf{Baseline} combines two fundamental training approaches, \emph{e.g.}, classification training and pairwise training for supervision. Since ZS-SBIR aims to retrieve photos according to the sketch query, the pairwise training uses the sketch as the anchor and selects the furthest positive and closest negative photos to construct a cross-modality triplet. % adopts a cross-modality PK sampler to construct training batches. It 
		%For each anchor sketch (photo), we select the furthest positive and closest negative photos (sketches) as a triplet to compute the embedding loss. 
		Based on this baseline, \textbf{MATHM} supplements the cross-modality triplet with two other forms of triplet (within-modality triplet and hybrid triplet), as well as a gradient-based weighting module. Learning from the within-modality triplet enhances the category discrimination under every single modality while learning from the hybrid triplet directly suppresses the modality gap. All three types of triplet loss are dynamically aggregated with the gradient-based weighting module to calculate the overall embedding loss. % Best viewed in color.
	}
	\vspace{-0.1cm} 
	\label{Fig:network}
\end{figure*}

%------------------------------------------------------------------------
\section{Related Works}

\subsection {Deep Metric Learning}
% The goal of Deep Metric Learning is 
Deep Metric Learning aims to map the input images from the pixel space to the embedding space with a deep Convolutional Neural Network (CNN), where images of the same class are close, and images of different classes are far away from each other.
Typically, an embedding loss or a classification loss is adopted to manipulate the features in the embedding space to achieve good within-class compactness and the between-class discrepancy.
%The typical approach to this goal is using an embedding loss or a classification loss to manipulate the features in the embedding space. % make the learned feature more discriminative.  
% The embedding loss usually focuses on mining the relationship among image pairs or triplet, which enforces a margin between positive and negative image pairs, 
The embedding loss usually focuses on mining the relationship among image pairs or triplets by enforcing a margin between positive and negative pairs in the embedding space, \emph{e.g.} Contrastive Loss \cite{hadsell2006dimensionality}, Triplet Loss \cite{weinberger2006distance, hermans2017defense, schroff2015facenet}, Lifted-Structure loss \cite{oh2016deep}, Angular loss \cite{wang2017deep}, N-Pair Loss \cite{sohn2016improved}, Multi-Similarity Loss \cite{wang2019multi}, and  FastAP \cite{cakir2019deep}.
And the classification loss utilizes a weight matrix as learnable class centers to exploit the global relationship between each sample to all categories, \emph{e.g.} Softmax Loss \cite{taigman2014deepface, parkhi2015deep}, NormFace \cite{wang2017normface}, SphereFace \cite{liu2017sphereface}, CosFace \cite{wang2018additive, wang2018cosface}, ArcFace \cite{deng2019arcface}, SoftTriplet Loss \cite{qian2019softtriple}, ProxyNCA \cite{movshovitz2017no}, and \emph{.etc}. 

%For the ZS-SBIR task, each category contains at least one cluster for each sketch and photo modality, thus the data distribution is multi-modal. 

As each category contains both sketch and photo images, the data distribution for the SBIR task is multi-modal. 
While most classification losses require that the data distribution is uni-modal (except for the SoftTriplet Loss, which uses multiple class centers), we select the more flexible triplet embedding loss to train our model, for it puts no restriction on the prior data distribution. 

%由于每个类别都包含用于SBIR任务的素描图像和照片图像，因此数据分布是多模式的。由于大部分分类损失要求数据分布是单峰的（（SoftTriplet损失除外，后者使用多个分类中心）。因此，我们选择更灵活的三重态嵌入损失来训练我们的模型，因为它对先前的数据分配没有任何限制 这些方法端到端训练CNN，以通过CNN将草图和照片图像映射到共享的嵌入空间中

\subsection {SBIR and ZS-SBIR}
The main challenge of the SBIR task is the modality discrepancy between hand-drawn sketches and real photos. Earlier works often use edge maps \cite{zhang2016sketch} extracted from real images as mid-level representation to bridge the gap between photos and sketches. Some specially designed feature descriptors, \emph{e.g.} Bag Of Words (BOG) \cite{eitz2010evaluation, eitz2010sketch}, HOG \cite{hu2013performance}, Learned KeyShapes (LKS) \cite{saavedra2015sketch}, are used to represent the sketches and edge-maps simultaneously. In recent years, many deep learning-based methods have emerged \cite{liu2017deep, sangkloy2016sketchy, lu2018learning, yu2017sketch, zhang2016sketchnet, zhang2018generative, yu2016sketch}. These methods train a CNN end-to-end to map the sketches and photo images into a shared embedding space. 
%These methods train a CNN to map the sketches and photo images into a shared embedding space in an end-to-end manner. 

%However, traditional SBIR methods generalize poorly on unseen categories, 
However, traditional SBIR methods assume that the training and testing set share the same categories, 
which is inapplicable to real-life scenarios. \cite{shen2018zero} proposes a more realistic setting named Zero-Shot Sketch-Based Image Retrieval (ZS-SBIR), where the testing classes are unseen during training. A majority of works on ZS-SBIR leverage semantic information obtained from language models \cite{mikolov2013efficient, pennington2014glove, miller1995wordnet, jiang1997semantic} to encourage knowledge transfer to unseen classes. 
To achieve this, \cite{pandey2020stacked} uses a stacked adversarial network to align features from different modalities; \cite{dey2019doodle, zhu2020ocean,yelamarthi2018zero, yang2016zero} combine the generative adversarial network with the auto-encoder to encode the visual feature and semantic feature to common space; \cite{shen2018zero, zhang2020zero} utilize graph neural network to fuse the semantic information with visual representation.
%, \cite{dutta2019semantically} use a semantic reconstruction loss to map the visual feature to semantic space
%\renewcommand{\figurewidth}{6cm}

%------------------------------------------------------------------------
\section{Methodology}

In this section, we start with introducing our metric learning baseline for ZS-SBIR, then describe our Modality-Aware Triplet Hard Mining strategy and the gradient-based weighting scheme.
% This section introduces our metric learning baseline for ZS-SBIR and then describes our Modality-Aware Hard Triplet Mining and gradient-based weighting scheme.
% then the auxiliary loss functions and the Complementary Cross-modal Feature Learning Network.

%-------------------------------------------------------------------------
\subsection{The Metric Learning Baseline} %
\subsubsection{Two Fundamental Training Approaches} %

%We slightly modified the common metric learning setting in \cite{musgrave2020metric} to form our metric learning baseline for ZS-SBIR. 
%is to adapt to the ZS-SBIR task. 因此我们可以从全局的层面来优化模型 分类loss的主要作用是稳定训练的过程, for which it's hard to find a global optimum

Let $X=\{(x_i^m, y_i)|y_i \in C, m \in \{s, p\} \}_{i=1}^N$ be the training set ($s$ denotes sketch while $p$ denotes photo), where $x_i^m$ is an image with label $y_i$ from modality $m$ and $C$ is the training class set. The distance function $D(x_i^{m_1}, x_j^{m_2})=||f_{\theta}(x_i^{m_1}) - f_{\theta}(x_j^{m_2})||_2$ calculates the Euclidean distance between an image pair. 
$f_{\theta}(\cdot)$ is the embedding function that is parameterized by ${\theta}$, which maps the images into a L2-normalized embedding space. % The output features are. 

Following the common practice in deep metric learning, a triplet embedding loss is combined with a softmax cross-entropy classification loss to form our baseline. % \footnote {The primary duty of the softmax cross-entropy classification loss is to stabilize the training process. Since it uses a weight matrix to store all categories' proxy, we can optimize our model from a global level. In contrast, the triplet embedding loss focuses on the relationship among a small subset of classes within the current mini-batch. Thus our model will likely be trapped in a bad local optimum if trained without the classification loss.}, 
The learning objective is formulated as:

\vspace{-0.2cm}
\begin{equation}
\begin{aligned}
\mathcal L_{all}=\mathcal L_{cls}+\lambda \mathcal L_{cross},
\label{equ:all}
\end{aligned}
\vspace{-0.0cm}
\end{equation}
where $\mathcal L_{cls}$ is the classification loss, and $\mathcal L_{cross}$ is the embedding loss. $\lambda$ is a hyper-parameter that controls the relative weight of the two kinds of losses. An overview of our baseline is shown in Figure \ref{Fig:network}. 

%\paragraph {Classification loss} is computed using the softmax cross-entropy loss function:
\paragraph {Classification Training.} 
The primary target of the classification training in this task is to stabilize the training process. We can optimize our model from a global level with a classification loss since it uses a weight matrix to store the proxies for all categories. In contrast, the pairwise training focuses on the relationship among a small subset of classes within the current mini-batch. Thus our model will likely be trapped in a bad local optimum if trained without the classification loss.
The standard softmax cross-entropy loss is adopted for classification training, which is defined as: 

\begin{equation}
\begin{aligned}
\mathcal L_{cls} = -\log{\frac{e^{W_{y_i}^T\cdot f(x_i^m)}}{\sum_{c \in C}e^{W_{c}^T\cdot f(x_i^m)}}} ,
\label{equ:cls}
\end{aligned}
\end{equation}
where $W_c$ is the weight term in the classifier $\Theta_C$ corresponding to category $c$. 
%where the input image $x_i^m$ can be either a sketch or a photo.
%To obtain a symmetric embedding about the origin of coordinate axis, the bias term of the classifier is removed following \cite{luo2019bag}.
%and the bias term of the classifier is removed following \cite{luo2019bag} to obtain a symmetric embedding about the	origin of the coordinate axis. 

\label{sec:definition}
% \paragraph{Embedding loss.} 
\paragraph{Pairwise Training.} 	
%, thus the relationship between images across modalities is of vital importance to train a strong ZS-SBIR baseline. 
Distinct from the traditional single modality image retrieval problems, the query sketches and the gallery images are from different modalities in the ZS-SBIR task.
We expect that for any anchor sketch image $x_a^s$, its positive photo image $x_i^p$ should be closer to $x_a^s$ than its negative photo image $x_j^p$ in the embedding space: 

\begin{equation}
\begin{aligned}
D(x_a^s, x_i^p)<D(x_a^s, x_j^p),y_a=y_i \ne y_j .
\label{equ:goal}
\end{aligned}
\end{equation}

%To achieve the above goal, we design a basic cross-modality triplet loss by mining the cross-modal triplet to reduce $D(x_a^s, x_i^p)$ and increase $D(x_a^s, x_i^p)$ directly.	
To achieve the above goal, we design a cross-modality triplet loss to directly reduce $D(x_a^s, x_i^p)$ and increase $D(x_a^s, x_i^p)$. One of the key issues of SBIR is to resolve the discrepancy between sketch and photo. Thus, it is vital to maintain a consistency scheme: if the photo closest to sketch $s$ is $p$, we hope that the sketch closest to photo $p$ is also $s$. So both sketch and photo are selected as anchors to form cross-modality triplets. The cross-modality triplet loss is then defined as:
% In spirit of symmetry, we utilizes both sketch and photo as the anchor images to form cross-modal triplets.  % the two kind of triplet relationships by using
%In spirit of symmetry, there should be similar constraint for an anchor image from the photo domain. 
%Following the above goal, the basic cross-modality triplet loss can be naturally defined as: 缓解差异 假如距离草图a最近的照片是b，那么我们希望距离照片b最近的草图也是a
%Thus the embedding loss of our baseline is defined as:

\vspace{-0.1cm}
\begin{equation}
\begin{aligned}
\mathcal L_{cross}&=[(D(x_a^s, x_i^p))-D(x_a^s, x_j^p)+\alpha]_+  \\
&+[(D(x_a^p, x_i^s))-D(x_a^p, x_j^s)+\alpha]_+  ,\\
\label{equ:L_basic}
\end{aligned}	\vspace{-0.5cm} 
\end{equation}
where the $x_a$ and $x_i$ are from the same category, while the $x_a$ and $x_j$ are from different categories.
A positive margin parameter $\alpha$ is added for better separability in the embedding space for different categories.

\subsubsection{Sampling Strategy} 
Following the conventional PK sampling strategy \cite{hermans2017defense}, we form a mini-batch by randomly sample $P$ classes and then sample $K$ images of each modality from each category. Thus the total batch size is $2\times PK$. For each sample in the mini-batch, we select the hardest positive and hardest negative from the corresponding modality within the mini-batch to form a triplet $T$:

\vspace{-0.2cm}
\begin{equation}
\begin{aligned}
\left\{
\begin{array}{lr}
T=(x_a^{m_a}, x_{pos}^{m_p}, x_{neg}^{m_n}) &  \\
pos=\arg\max_{i}{D(x_a^{m_a}, x_i^{m_p})}, & y_a = y_{i} \\
neg=\arg\min_{j}{D(x_a^{m_a}, x_j^{m_n})}, & y_a \neq y_{j}  
\end{array}
\right.
\label{equ:BH}
\end{aligned}
\end{equation}	
where $m_a$, $m_p$, $m_n$ stand for the modality of the anchor, positive, and negative images, respectively. 
For our cross-modality triplet loss, we have $m_a\neq m_p = m_n$, which means that we only select positive and negative images with different modalities from the anchor image to form a triplet. % the modality of the anchor images is different from that of its positive and negative images. %, thus .
% In Eq. (\ref{equ:\mathcal L_basic}), for example, when the anchor image is from the sketch domain, the modality of the positive and negative images should be photo, and vice versa.

\subsubsection {Network Architecture} 
%To construct a simple yet strong ZS-SBIR framework, 
We adopt a single stream CNN as the feature extractor to map images from different modalities to a shared embedding space. Following \cite{liu2019semantic}, we select CSE-ResNet50 \cite{lu2018learning} as our backbone model. The CSE-ResNet50 integrates the conditional SE module into ResBlocks \cite{he2016deep}, which effectively handles inputs from different modalities.

%-------------------------------------------------------------------------
\subsection{Modality-Aware Triplet Hard Mining}	% Auxiliary learning objective
\label{sec:MATHM}

%introduce the two extra learning objectives by introduce the decomposition of distance.
\begin{figure}[t]
% 	\vspace{-0.2cm}
	
	\centering
	\begin{subfigure}{0.4\linewidth}
		\includegraphics[width=\linewidth]{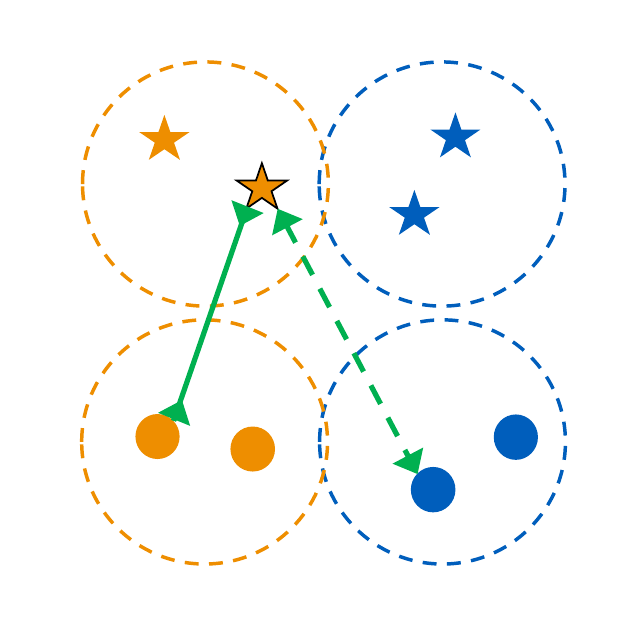}
		\caption{Baseline}
		\label{fig:gap}
	\end{subfigure}
	\quad
	\begin{subfigure}{0.4\linewidth}
		\includegraphics[width=\linewidth]{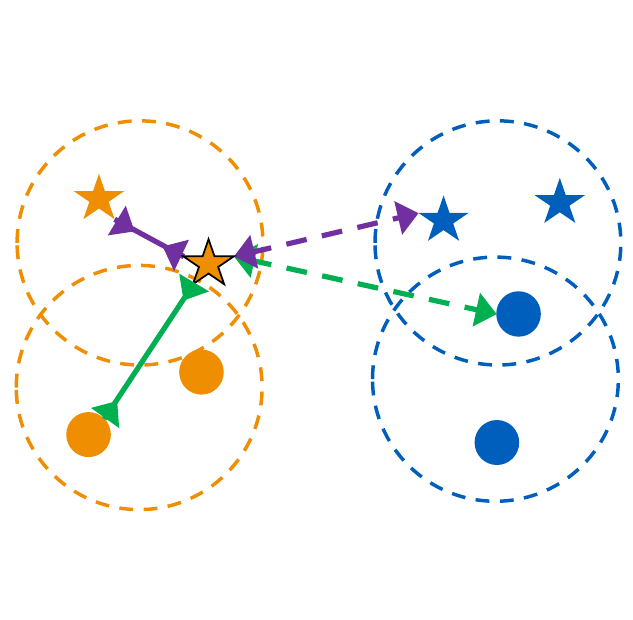}
		\caption{MATHM}
		\label{fig:good}
	\end{subfigure}
%	\subfigure[Baseline]{
%		\includegraphics[width=\subfigwidth]{Figures/relation-gap.pdf}
%	}
%	\quad 
%	\subfigure[MATHM]{
%		\includegraphics[width=\subfigwidth]{Figures/relation-good.pdf}
%	}
	\vspace{-0.2cm} 
	\caption{
		(a) Our baseline has high tolerance on the modality gap, where the gap hurts within-class compactness in the testing phase and deteriorates the model's performance for the ZS-SBIR task. (b) In order to obtain better within-class compactness, we not only impose baseline constraints but also explicitly suppress the modality gap with the MATHM. Each color represents a particular class. The stars denote sketches, and circles denote photos. Best viewed in color.
	}
	\vspace{-0.3cm} 
	\label{Fig:relation}
\end{figure}
%Mining the cross-modal triplet relationship is not enough to train a good feature extractor.

Our Modality-Aware Triplet Hard Mining aims at utilizing hard triplets of different modal compositions to train a robust ZS-SBIR model. 

For the SBIR problem, the distance between the query sketches and gallery photos is related to both the class level discrepancy and the modality gap. % closely related to two properties of the embedding space: 1) the between-class discrepancy and 2) the modality gap.
Assuming that the distance between two images $d_{i, j}=D(x_i^{m_i}, x_j^{m_j})$ can be decomposed into two parts: 
% a class-specific term $d_{i, j}^c$ and a modality-specific term $d_{i, j}^m$ as:

\vspace{-0.2cm}
\begin{equation}
\begin{aligned}
d_{i, j} = G(d_{i, j}^c, d_{i, j}^m) ,
\label{equ:decompose}
\end{aligned}
\end{equation}
where $G$ is an abstract function that aggregates the class-specific distance $d_{i, j}^c$ and the modality-specific distance $d_{i, j}^m$. 
The cross-modality triplet loss optimizes the overall distance $d_{i, j}$ on cross-modality triplets, which neither explicitly reduces the modality gap nor ensures that the within-class distance is smaller than the between-class distance. 
As the example shown in Figure \ref{Fig:relation}, although the cross-modality triplet loss has been satisfied, the modality gap and within-class variation are still large.
To train a strong cross-modality embedding function, we need to reduce the modality gap and make each testing class separable from the other.
Therefore, we extend our baseline with two other triplet relationships, \emph{e.g.}, the within-modality triplet relationship and hybrid triplet relationship.
The standard triplet embedding loss with batch-hard triplet mining (Eq. (\ref{equ:BH})) is used for loss calculation, detailed as follows. 
% Intra-modal Triplet loss To learning a discriminative embedding for both sketch and photo inputs,
\paragraph {Within-Modality Triplet.}  
This is a typical triplet relationship in single modal metric learning tasks. The positive and negative images with the same modality as the anchor image ($m_a = m_n = m_p$) are selected to form a triplet for learning discriminative embedding within each modality. The within-modal triplet loss is defined as:
\vspace{-0.0cm} 
\begin{equation}
\begin{aligned}
\mathcal L_{in}&=[(D(x_a^s, x_i^s))-D(x_a^s, x_j^s)+\alpha]_+ \\
&+[(D(x_a^p, x_i^p))-D(x_a^p, x_j^p)+\alpha]_+ .\\
\vspace{-0.2cm} \label{equ:L_intra}
\end{aligned}	\vspace{-0.5cm} 
\end{equation}
The above loss function encourages our model to learn a class discriminative feature embedding for input sketches and photos. Since no cross-modality relationship is included, using the within-modality triplet loss alone is less effective to handle this cross-modality matching problem. % train a ZS-SBIR model.

% Cross-modality Triplet loss 他很少被大家研究
% means that the modality of the anchor image is the same as the negative image but different from the positive image ($m_a = m_n \neq m_p$). 
\paragraph{Hybrid Triplet.} % {Partially Cross-Modality Triplet} 
% Most SBIR datasets do not provide paired sketch and photo images, it's hard to directly reduce the modality gap by align the feature extracted from sketch or photo.
Modality gap is a major concern in the cross-modality matching tasks. One of the common solutions to this problem is to align the feature distribution of two modalities using the Generative Adversarial Mechanism \cite{GAN}. The learned feature embedding for sketches and photos is expected to be similar enough to ``fool'' the discriminator, thus the modality gap is minimized.
However, reducing the modality gap is not the only challenge for the SBIR problem, learning discriminative feature embedding is also of vital importance. Due to the large inherent differences between sketches and photos, directly narrowing the modality gap might weaken the expressiveness of feature embedding and further compromise the performance of SBIR. % . Leading the model to over-fit on training classes
%草图和照片的风格和形式差异非常大
%Due to the lack of the paired sketch and photo images, it's hard to directly align the feature extracted from sketch or photo. To tackle this problem, one of the most popular strategy is the adversary training. By alternately updating the encoder and decoder with opposite goal, the modality gap could be minimized. 
% However, this strategy only focuses on reducing the modality gap, ignoring the fact that the feature distribution of two modality is naturally different. If we push the model too hard to align the feature distribution of two modality on training set, it is likely to over-fit and hurt the between class discrepancy of the embedding space. Thus the performance of SBIR is compromised. 
Thus we design the hybrid triplet loss to tackle the modality discrepancy problem while preserving the between-class discrepancy.
%In order to tackle the modality discrepancy without affecting the between class discrepancy, we design the hybrid triplets loss to handle the contradiction between the above two issue.

The hybrid triplet is composed of an anchor image, a cross-modality positive image, and a same-modality negative image ($m_a = m_n \neq m_p$). The corresponding loss function is written as: % It is rarely studied by previous works.

\begin{equation}
\begin{aligned}
\mathcal L_{hyb}&=[(D(x_a^s, x_i^p))-D(x_a^s, x_j^s)+\alpha]_+  \\
&+[(D(x_a^p, x_i^s))-D(x_a^p, x_j^p)+\alpha]_+ . \\
\label{equ:L_cross}
\end{aligned}	\vspace{-0.5cm} 
\end{equation}
The above loss function aims to minimize the distance between cross-modality positives and maximizes that of the same modality negatives. The pull term minimizes the modality gap $d_{a, i}^m$ and the within-class variation $d_{a, i}^c$ concurrently, while the push term maximize the between-class discrepancy. % Learning with $L_{hyb}$ is beneficial to both of the above goal.
% Since the within-modal triplet loss has already minimized the within-class variation $d_{a, i}^c$ and maximized the distance between the same modal negatives, the above loss can only be further minimized by reducing the modality gap $d_{a, i}^m$. Therefore, the primary effect of hybrid triplet loss is suppressing the modality gap while maintaining the between class discrepancy. We will experimentally validate this assumption in Sec \ref{Sec:adv-exp}. 
Therefore, the hybrid triplet loss is capable of suppressing the modality gap while maintaining the between-class discrepancy. We will experimentally validate this assumption in Section \ref{Sec:adv-exp}.

%%%%%%%%%%% new
%Except for the Partially Cross-Modality Triplet loss, there are many other ways to narrow the modality gap, where the most popular one is the adversary training. By alternately updating the encoder and decoder with opposite goal, the modality gap could be minimized. However, this strategy only focuses on reducing the modality gap, ignoring the fact that the feature distribution of two modality is naturally different. If we push the model too hard to align the feature distribution of two modality on training set, it is likely to over-fit and hurt the between class discrepancy of the embedding space. Thus the performance of SBIR is compromised. 
% In comparison, MATHM achieves this goal in a softer manner.

% From the analysis above, none of the three triplet relationships is sufficient to train a robust model, but each of them is indispensable. Hence it's imperative to combine all these triplet relationships to form the overall objective of MATHM. 

%\renewcommand{\figurewidth}{6cm}

%-------------------------------------------------------------------------
\subsection{Gradient-based Weighting Scheme}	% Dynamic weighting strategy
% 增加梯度随epoch变化的图 在设计损失时，考虑他们产生的梯度至关重要

\begin{figure}[t]
	\vspace{-0.1cm}
	\centering		
	\small
	\includegraphics[width=0.8\linewidth]{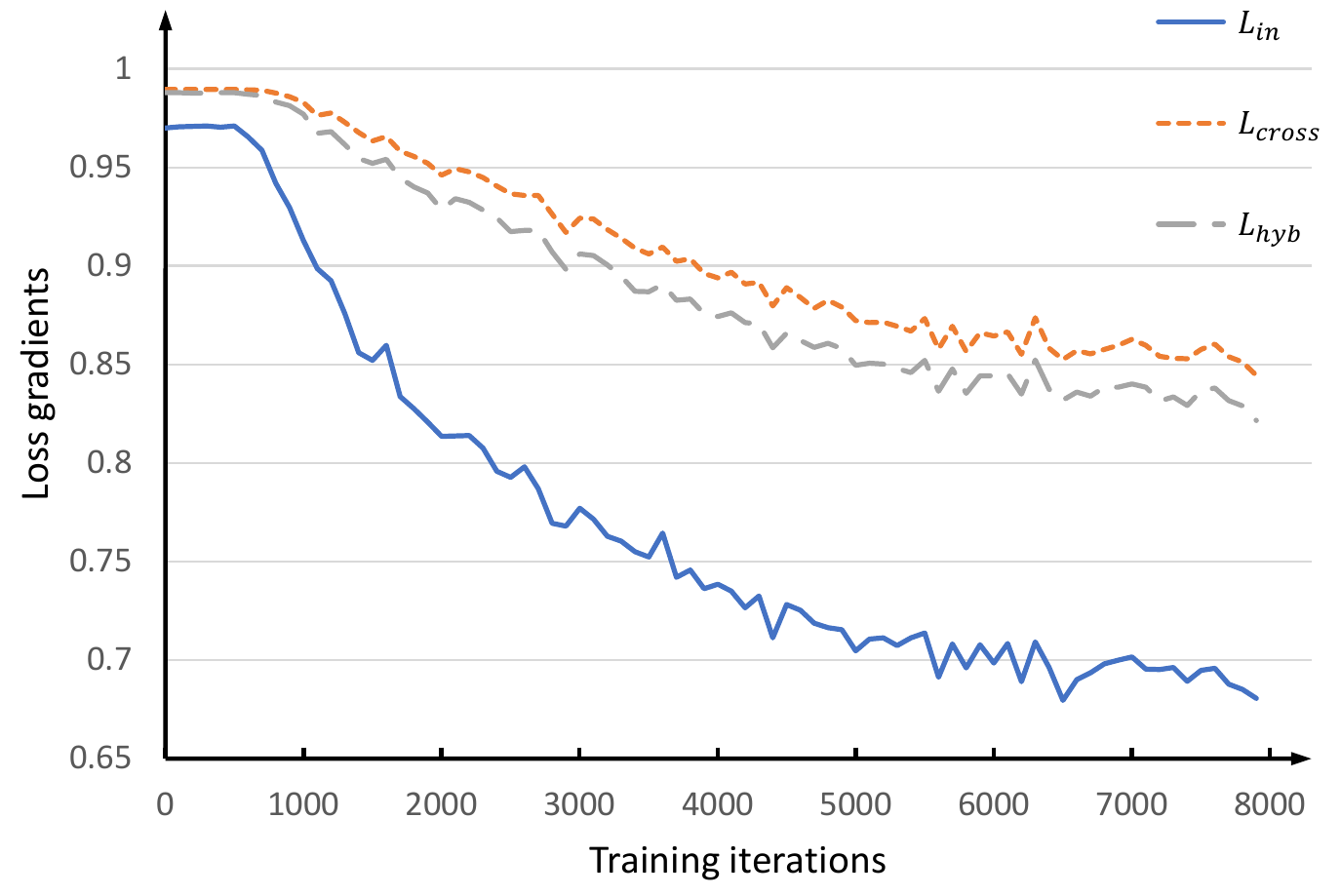}
	\vspace{-0.0cm} \caption{The changes in gradient during training. % Here, we use a summation to aggregate the losses.
	}
	\vspace{-0.2cm} \label{Fig:grad}
\end{figure}
We will first briefly revisit the gradient calculation of triplet loss and then introduce our gradient-based weighting scheme for aggregating different triplet losses.
%As the above three functions are built on triplet loss:表达式 内在的差异 internal 
%To calculate the gradient \emph{w.r.t.} the above loss functions, we first formulate
%\paragraph {Gradient Calculation.} 
%As the gradient-based optimization methods have dominated deep CNN training, it's vital to investigate the gradient produced by each loss. 
%We formulate the general form of triplet embedding loss as:
The general form of triplet embedding loss is formulated as:
%The general form of triplet loss embedding is defined as:
\begin{equation}
\begin{aligned}
\mathcal L_{i}&=\frac{1}{N}\sum[d_{ap}-d_{an}+\alpha]_+  ,
\end{aligned}
\end{equation}
where $[\bullet +\alpha]_+$ is the hinge function. % Therefore,
The above equation implies that only the triplets with a positive value in $(d_{ap}-d_{an}+\alpha)$ will produce gradients for optimization. The gradient $g_i$ \emph{w.r.t.} $\mathcal L_i$ is calculated as:

\begin{equation}
\begin{aligned}
g_i&=\frac{1}{N}\sum sgn([ d_{ap}-d_{an}+\alpha]_+ ) .
\end{aligned}
\end{equation}
%in which $sgn(x)$ is the sign function that returns $1$ when $x$ is greater than zero, and returns $0$ when $x$ equals to zero.

\begin{table*}[t]
	\renewcommand\arraystretch{1.1}
	\footnotesize
	\centering		
	\caption{The performance comparison of MATHM and existing methods on ZS-SBIR tasks. ``$\dagger$'' denotes experiments using binary hashing codes. The rest use real-valued features. ITQ \cite{gong2012iterative} is used to binarize our feature to compare with the zero-shot hashing methods. The best results are in \textbf{bold}, while\ the best results of previous methods are in \color{blue}{blue}.
		%	``-'' indicates the results are not presented by the authors on that metric.
	} \vspace{-0.1cm}
	\begin{tabular}{l|lccccccc}
		\hline
		\multirow{2}{*}{Task} &\multirow{2}{*}{Method}   & \multirow{2}{*}{dim}  & \multicolumn{2}{c}{TU-Berlin Ext.}                    & \multicolumn{2}{c}{Sketchy Ext.}                & \multicolumn{2}{c}{Sketchy Ext.(\cite{yelamarthi2018zero} Split) }               \\ \cline{4-9} 
		& &      & mAP@all             & Prec@100            & mAP@all             & Prec@100            & mAP@200             & Prec@200            \\ \hline
		\multirow{2}{*}{SBIR} &GN-triplet \cite{sangkloy2016sketchy}                  & 1024        & 18.9 & 24.1 & 21.1 & 31.0   & 8.3  & 16.9 \\
		&DSH \cite{liu2017deep}                                 & 64$\dagger$ & 12.2 & 19.8 & 16.4 & 22.7 & 5.9  & 15.3 \\ \hline
		
		\multirow{2}{*}{ZSL} &SAE \cite{kodirov2017semantic}                         & 300         & 16.1 & 21.0   & 21.0   & 30.2 & 13.6 & 23.8 \\
		&ZSH \cite{yang2016zero}                                & 64          & 13.9 & 17.4 & 16.5 & 21.7 & -    & -    \\ \hline
		
		\multirow{20}{*}{ZS-SBIR}& ZSIH \cite{shen2018zero}                               & 64          & 22.0   & 29.1 & 25.4 & 34.0   & -    & -    \\
		& \multirow{2}{*}{EMS \cite{lu2018learning}}             & 512         & 25.9 & 36.9 & -    & -    & -    & -    \\
		& & 64          & 16.5 & 25.2 & -    & -    & -    & -    \\
		& CCAE \cite{yelamarthi2018zero}                         & 4096        & -    & -    & 19.6 & 28.4 & 15.6 & 26.0   \\
		%& CVAE \cite{yelamarthi2018zero}                         & 4096        & -    & -    & -    & -    & 22.5 & 33.3 \\
		& \multirow{2}{*}{SEM-PCYC \cite{dutta2019semantically}} & 64          & 29.7 & 42.6 & 34.9 & 46.3 & -    & -    \\
		& & 64$\dagger$ & 29.3 & 39.2 & 34.4 & 39.9 & -    & -    \\
		& \multirow{2}{*}{SAKE \cite{liu2019semantic}}           & 512         & \color{blue}{47.5} & \color{blue}{59.9} & \color{blue}{54.7} & \color{blue}{69.2} & \color{blue}{49.7} & \color{blue}{59.8} \\
		& & 64$\dagger$ & \color{blue}{35.9} & \color{blue}{48.1} & 36.4 & 48.7 & \color{blue}{35.6} & \color{blue}{47.7} \\
		& SketchGCN \cite{zhang2020zero}                         & 1024        & 32.4 & 50.5 & 38.2 & 53.8 & -    & -    \\
		& Doodle2Search \cite{dey2019doodle}                     & 4096        & 10.9 & -    & 36.9 & -    & -    & -    \\
		& STYLE-GUIDE \cite{dutta2019style}                      & 4096        & 25.4 & 35.6 & 37.6 & 48.4 & 35.8 & 40.0   \\
		& BDT \cite{li2019bi}                                    & 1024        & -    & -    & -    & -    & 28.1 & 39.4 \\
		& OCEAN \cite{zhu2020ocean}                              & 512         & 33.3 & 46.7 & 46.2 & 59.0   & -    & -    \\
		& \multirow{2}{*}{PCMSN \cite{deng2020progressive}}      & 64          & 42.4 & 51.7 & 52.3 & 61.6 & -    & -    \\
		& & 64$\dagger$ & 35.5 & 45.2 & \color{blue}{50.6} & \color{blue}{61.5} & -    & -    \\ % \cline{2-9} 
		& \multirow{2}{*}{Baseline (ours)}     & 512   & 47.88 $\pm$ 0.07          & 58.28 $\pm$ 0.14          & 58.72   $\pm$ 0.42        & 70.26   $\pm$ 0.42        & 48.78   $\pm$ 0.21        & 60.06   $\pm$ 0.25 \\
		& & 64$\dagger$  & 39.76 $\pm$ 0.28 & 50.72 $\pm$ 0.12 & 49.72 $\pm$ 0.43 & 63.28 $\pm$ 0.28 & 45.08 $\pm$ 0.36 & 54.32 $\pm$ 0.31 \\ 
		& \multirow{2}{*}{MATHM (ours)}     & 512  & \textbf{50.82 $\pm$ 0.12} & \textbf{60.62 $\pm$ 0.18} & \textbf{62.92 $\pm$ 0.14} & \textbf{73.80 $\pm$ 0.11} & \textbf{52.26 $\pm$ 0.11} & \textbf{63.28 $\pm$ 0.20} \\
		& & 64$\dagger$  & \textbf{42.94 $\pm$ 0.58} & \textbf{54.06 $\pm$ 0.47} & \textbf{55.36 $\pm$ 0.60}  & \textbf{66.00 $\pm$ 0.43} & \textbf{47.78 $\pm$ 0.57} & \textbf{57.58 $\pm$ 0.46} \\ \hline
	\end{tabular}
	\\
	\label{tab:compare}\vspace{-0.1cm}
\end{table*}
% \paragraph {Gradient-Based Weighting Scheme.} 
As each of the three triplet losses plays an essential role in learning a good embedding, we prefer that they could contribute equally in the training stage.
One typical way to integrate multiple loss functions is by summation, but it is not optimal. The contribution of each loss function depends on the gradient they produced during the backward pass. % is measured by the gradient they produced. 
But the gradient of each triplet loss decays at varying speeds due to their different convergence targets (as Figure \ref{Fig:grad} shows). Some of them are hard to converge, for which they consistently produce strong gradients. This phenomenon indicates that the harder ones will dominate the optimization if we aggregate these losses by summation. 
Thus we design a gradient-based weighting scheme to make equal use of all these losses. Assume the values of the above loss functions (Eq. (\ref{equ:L_basic}, \ref{equ:L_intra}, \ref{equ:L_cross})) at the current iteration are $\mathcal L_i (i=1,2,3)$, where the gradient \emph{w.r.t.} $\mathcal L_i$ is $g_i$, and the weight of $\mathcal L_i$ is $w_i$.
We gather the three components of MATHM with the gradient-based weighting scheme to form the overall embedding loss:
%The overall embedding loss of MATHM gathered with the gradient-based weighting strategy is defined as:
\vspace{-0.2cm}
\begin{equation}
\begin{aligned}
\mathcal L_{\text{MATHM}}&=\sum_{i=1}^n{w_i \mathcal L_i} .
%\mathcal L_{embed}&=\sum_{i=1}^n{w_i \mathcal L_i} % & \text { (n=3)} 
\label{equ:L_metric}
\end{aligned}
\end{equation}
The $\mathcal L_{cross}$ in Eq. (\ref{equ:all}) is replaced with $\mathcal L_{\text {MATHM}}$ to form our total learning objective.

%Our goal is to keep the gradient produced by each loss function at every iteration to be equal while the total magnitude of gradient remains the same. 
% Our goal is to keep the gradient produced by each loss function at every iteration to be equal.
Our goal is to ensure that each loss function produces an equal amount of gradient. Meanwhile, the total magnitude of gradients should remain the same after weighting. This objective can be formulated as:

\vspace{-0.0cm}
\begin{equation}
\begin{aligned}
\left\{
\begin{array}{lr}
w_i g_i = w_j g_j ,\quad \forall i \neq j   \vspace{1ex}  \\
\sum_{i=1}^n{w_i g_i} = \sum_{i=1}^n{g_i}
\end{array} 
\right.
\label{equ:equation}
\end{aligned}
\end{equation}
We solve the above equation set to calculate $w_i$:
%Then, $w_i$ is calculated by solving the above equation set:	
\vspace{-0.0cm}
\begin{equation}
\begin{aligned}
%w_i&= \frac{1}{n g_i}\sum_{k=1}^n{g_k}
%w_i&= \frac{\sum_{k=1}^n{g_k}}{n g_i}\\
%w_i&= \frac{1}{g_i}\frac{\sum_{k=1}^n{g_k}}{n }\\
%w_i&= \frac{1}{n}\frac{\sum_{k=1}^n{g_k}}{g_i}\\
w_i&= \frac{1}{n}\sum_{k=1}^n{\frac{g_k}{g_i}} ,\\
%w_i&= \frac{1}{n}\sum_{k=1}^n{}{\frac{g_k}{\max(g_i, eps)}} % & \text { (n=3)}
\label{equ:weight}
\end{aligned} \vspace{-0.5cm}
\end{equation}
where $n$ equals to $3$ in this case. % and $eps$ is empirically set to 0.1 for numerical stability. 

The gradient-based weighting scheme ensures that each loss function produces an equal magnitude of gradients in every step, reasonably aggregating the three triplet losses. % Algorithm \ref{alg:all} shows the overall training procedure of MATHM. % The overall training procedure is shown in Algorithm \ref{alg:all}.

%\begin{algorithm}[t] 
%	\renewcommand{\algorithmicrequire}{\textbf{Input:}}
%	\renewcommand{\algorithmicensure}{\textbf{Output:}}
%	\caption{Training Procedure of the proposed MATHM} 
%	\label{alg:all} 
%	\begin{algorithmic}[1] 
%		\Require
%		Dataset $X=\{(x_i^m, y_i)|y_i \in C, m \in \{s, p\} \}_{i=1}^N$;
%		Sampling parameter $P$, $K$; 
%		Max training iteration $T$;
%		Hyper-parameter $\lambda$.
%		\Ensure			
%		Embedding function $f_{\theta}(x)$ with parameter $\theta$; 
%		%			optimal $x^{*}$ 
%		\State Initialize $\theta$; 
%		\For {$i=1$ to $T$} 
%		\State Forward model to generate $f_{\theta}(x_i^{s})$, $f_{\theta}(x_i^{p})$; 
%		\State Calculate classification loss $\mathcal L_{cls}$ with Eq. (\ref{equ:cls}); 
%		\State Calculate cross-modality triplet loss $\mathcal L_{cross}$ with 
%		\Statex \ \ \ \ Eq. (\ref{equ:L_basic});
%		\State Calculate within-modality triplet loss $\mathcal L_{in}$ with Eq. (\ref{equ:L_intra}); 
%		\State Calculate hybrid triplet loss $\mathcal L_{hyb}$ with Eq. (\ref{equ:L_cross});
%		\State Calculate the overall embedding loss $\mathcal L_{MATHM}$ with 
%		\Statex \ \ \ \ the gradient-based weighting scheme according to 
%		\Statex \ \ \ \ Eq. (\ref{equ:L_metric});
%		\State Calculate the overall training loss $\mathcal L_{all}$ with Eq. (\ref{equ:all});
%		\State Update $\theta \stackrel{+}{\leftarrow} \nabla_{\theta}(\mathcal L_{all}) $;
%		
%		\EndFor \\
%		
%		\Return $\theta$
%	\end{algorithmic}
%\end{algorithm}
%------------------------------------------------------------------------
\section{Experimental Results}

\begin{table*}[h]
	\renewcommand\arraystretch{1.2}
	%	\footnotesize
	\centering
	\scriptsize
	\caption{
		The quantitative analysis of the proposed methods and the generative adversarial network on the TU-Berlin testing set. $\uparrow$ means the larger the better and $\downarrow$ means the smaller the better. The best results are in bold. The confidence intervals are omitted for clarity. % \textbf{GAN} stands for the generative adversarial network.  in this table
	}
	\vspace{-0.2cm}
	\begin{tabular}{c|c|cc|c|cc|cc}
		\hline
		\multirow{2}{*}{Methods}  & \multirow{2}{*}{Similarity} & \multirow{2}{*}{Within-Class} & \multirow{2}{*}{Between-Class} &               Within-Class                &                      \multicolumn{2}{c|}{Between-Class Discrepancy}                      &     \multirow{2}{*}{mAP@all$\uparrow$}     &    \multirow{2}{*}{Prec@100$\uparrow$}     \\ \cline{6-7}
		&                                    &                               &                                &        {Modality gap$\downarrow$}         & \multicolumn{1}{c|}{Same-Modality$\uparrow$} &         Cross-Modality$\uparrow$          &                                            &                                            \\ \hline
		\multirow{2}{*}{Baseline} &           Same-Modality            &        0.527         &        0.070          &     \multirow{2}{*}{0.259}      &  \multirow{2}{*}{\textbf{0.456}}   &     \multirow{2}{*}{0.224}      &     \multirow{2}{*}{47.88}      &     \multirow{2}{*}{58.28}      \\
		&           Cross-Modality           &        0.268         &        0.044          &                                           &                                              &                                           &                                            &                                            \\ \hline
		\multirow{2}{*}{MATHM}   &           Same-Modality            &        0.531         &        0.129          &     \multirow{2}{*}{0.158}      &       \multirow{2}{*}{0.402}       & \multirow{2}{*}{\textbf{0.247}} & \multirow{2}{*}{\textbf{50.82}} & \multirow{2}{*}{\textbf{60.62}} \\
		&           Cross-Modality           &        0.373         &        0.126          &                                           &                                              &                                           &                                            &                                            \\ \hline
		\multirow{2}{*}{GAN}    &           Same-Modality            &        0.653         &        0.368          & \multirow{2}{*}{\textbf{0.072}} &       \multirow{2}{*}{0.285}       &     \multirow{2}{*}{0.218}      &      \multirow{2}{*}{40.19}      &      \multirow{2}{*}{56.81}      \\
		&           Cross-Modality           &        0.581         &        0.363          &                                           &                                              &                                           &                                            &                                            \\ \hline
	\end{tabular}\label{Tab:sim-gap}
	\vspace{-0.2cm}
\end{table*}

%-------------------------------------------------------------------------
\subsection{Datasets and Setup} 
We evaluate our model on two popular large-scale SBIR benchmarks: the TU-Berlin \cite{eitz2012humans} Extended and the Sketchy \cite{sangkloy2016sketchy} Extended. 

\noindent {\bf TU-Berlin Extended} contains 20,000 sketch images of 250 categories from the TU-Berlin dataset and 204,489 extra photo images from \cite{zhang2016sketchnet}. 
Following \cite{shen2018zero}, we choose 30 identical classes as the unseen set while the other 220 categories are used for training. % from internet and Imagenet. 

\noindent {\bf Sketchy Extended} originally contains 12,500 photo images and 75,479 sketch images from 125 different classes. \cite{liu2017deep} extended this dataset by adding 60,502 natural images to the photo domain, so the total number of photo images is 73,002. \cite{shen2018zero} propose a zero-shot evaluation protocol by randomly chosen 25 categories as the unseen set. 
% To conduct experiments under the zero-shot setting, \cite{shen2018zero} propose an evaluation protocol by randomly chosen 25 categories as the unseen set for testing. 
But \cite{yelamarthi2018zero} suggests that some randomly chosen categories in the unseen set might have appeared in the ImageNet\cite{deng2009imagenet}, which violates the zero-shot assumption when using the ImageNet pre-trained weights for model initialization. Therefore they put forward a new protocol using 21 carefully selected categories orthogonal to the ImageNet classes as the unseen set.  
We will conduct our experiments under both of these two protocols to validate our method.
% For clearly, we refer to the partition in \cite{shen2018zero} and \cite{yelamarthi2018zero} as P-Rand and P-Zero respectively. 

%-------------------------------------------------------------------------
\subsection{Implementation details}

We implement our method using PyTorch \cite{paszke2017automatic} toolbox with one RTX2080Ti GPU. An ImageNet pre-trained SE-ResNet50 \cite{hu2018squeeze} is used to initialize our model. We use Adam \cite{kingma2014adam} optimizer with parameter $\beta_1 = 0.9$, $\beta_2 = 0.999$ to train our model. The margin parameter $\alpha$ in all our triplet losses is set to 0.2.
%The learning rate is set to 1e-5 for the CNN backbone and 1e-4 for other parameters. 
The initial learning rate is set to 1e-4, and cosinely \cite{loshchilov2016sgdr} decays to zero during training. For all the pre-trained layers, the learning rate is set to $0.1\times$ the base learning rate.
The batch size is set to 128, where the number of classes $P=16$ and the number of images per class $K=4$ for our sampler.
%containing P=16 classes and K=4 images for each modal. 
The whole network is trained for 8000 iterations to converge. We do five training runs in all our experiments and report the average of mean Average Precision (mAP@all) and Precision (Prec@100), as well as the 95\% confidence intervals.

%-------------------------------------------------------------------------
\subsection{Comparison with the State-of-the-Art}

We compare our methods with several existing works on ZS-SBIR. Most of them use semantic information. The earlier works such as ZSIH \cite{shen2018zero}, CCAE, CVAE \cite{yelamarthi2018zero}, Doodle2Search \cite{dey2019doodle} use a relatively simple framework, which adopts a single auto-encoder or reconstruction loss to combine semantic information with the visual feature. Recent works like OCEAN \cite{zhu2020ocean}, SketchGCN \cite{zhang2020zero}, PCMSN \cite{deng2020progressive} and SEM-PCYC \cite{dutta2019semantically} use more complicated structures. They combine multiple generative adversarial networks or graph neural networks with auto-encoder for learning a shared visual-semantic embedding space. Another recent work, SAKE \cite{liu2019semantic}, designs a teacher-student architecture for learning semantic information while preserving the pre-trained knowledge from ImageNet. We also compare our methods with another metric learning approach, EMS \cite{lu2018learning}, which enforces a class-level margin in the euclidean embedding space by using the Euclidean Margin Softmax (EMS) as the classification loss.

As Table \ref{tab:compare} shows, our simple baseline surpasses most of the existing methods. MATHM further improves the retrieval accuracy and outperforms all the state-of-the-art methods by a large margin. % in all three experiments. 	
Compared with one previous classification loss EMS, our triplet-based embedding losses show a significant performance boost. The competitive performance of our approach demonstrates that pairwise training is more effective than classification training under the cross-modality setting.

\begin{table*}[t]
	\renewcommand\arraystretch{1.1}
	\centering		
	\caption{Analysis on each component of the proposed method. %GW denotes the Gradient-based Weighting scheme.
	}
	\footnotesize
	\vspace{-0.2cm} 
	\begin{tabular}{cccccccccc}
		\hline
		\multirow{2}{*}{$\mathcal L_{cross}$} & \multirow{2}{*}{$\mathcal L_{in}$} & \multirow{2}{*}{$\mathcal L_{hyb}$} & \multirow{2}{*}{GW} & \multicolumn{2}{c}{TU-Berlin Ext.}    & \multicolumn{2}{c}{Sketchy Ext.}             & \multicolumn{2}{c}{Sketchy Ext. (\cite{yelamarthi2018zero} Split)}             \\ \cline{5-10} 
		&                        &                        &                     & mAP@all                   & Prec@100                  & mAP@all                   & Prec@100                  & mAP@200                   & Prec@200                  \\ \hline
		&                        &                        &                     & 46.00 $\pm$   0.16        & 56.42 $\pm$   0.18        & 54.40 $\pm$ 0.21          & 66.02   $\pm$ 0.27        & 45.94   $\pm$ 0.11        & 57.70   $\pm$ 0.18        \\
		$\checkmark$           &                        &                        &                     & 47.88 $\pm$ 0.07          & 58.28 $\pm$ 0.14          & 58.72   $\pm$ 0.42        & 70.26   $\pm$ 0.42        & 48.78   $\pm$ 0.21        & 60.06   $\pm$ 0.25        \\
		& $\checkmark$           &                        &                     & 47.58 $\pm$ 0.21          & 57.78 $\pm$ 0.13          & 57.44   $\pm$ 0.31        & 68.68   $\pm$ 0.31        & 48.58   $\pm$ 0.21        & 60.56   $\pm$ 0.29        \\
		&                        & $\checkmark$           &                     & 48.78 $\pm$ 0.27          & 58.88 $\pm$ 0.31          & 59.36   $\pm$ 0.39        & 70.96   $\pm$ 0.38        & 49.52   $\pm$ 0.07        & 61.06   $\pm$ 0.10        \\
		$\checkmark$           & $\checkmark$           &                        &                     & 48.97 $\pm$ 0.34          & 59.17 $\pm$ 0.22          & 60.50   $\pm$ 0.18        & 72.70   $\pm$ 0.19        & 51.16   $\pm$ 0.16        & 62.94   $\pm$ 0.21        \\
		$\checkmark$           &            & $\checkmark$                       &                     & 49.12 $\pm$ 0.16 & 59.36 $\pm$ 0.13 & 61.14 $\pm$ 0.30 & 72.98 $\pm$ 0.19 & 51.70 $\pm$ 0.13 & 63.50 $\pm$ 0.24        \\
		$\checkmark$           & $\checkmark$           & $\checkmark$           &                     & 50.40 $\pm$ 0.14          & 60.28 $\pm$ 0.08          & 62.20   $\pm$ 0.11        & 73.20   $\pm$ 0.07        & 51.84   $\pm$ 0.14        & 62.72   $\pm$ 0.12        \\
		$\checkmark$           & $\checkmark$           & $\checkmark$           & $\checkmark$        & \textbf{50.82 $\pm$ 0.12} & \textbf{60.62 $\pm$ 0.18} & \textbf{62.92 $\pm$ 0.14} & \textbf{73.80 $\pm$ 0.11} & \textbf{52.26 $\pm$ 0.11} & \textbf{63.28 $\pm$ 0.20} \\ \hline
	\end{tabular}
	\vspace{-0.2cm} 
	% {Performance comparison between different embedding loss functions and their combinations. In all these experiments, a cross-entropy loss is used as classification loss.}
	\label{tab:loss}\vspace{-0.2cm}
\end{table*}

%-------------------------------------------------------------------------
%\subsection{Quantitative Analysis}
%\paragraph{Comparison with other feature alignment methods.}
\subsection{Effect of MATHM}
\label{Sec:adv-exp}
\begin{figure}[t]
	\includegraphics[width=0.96\linewidth]{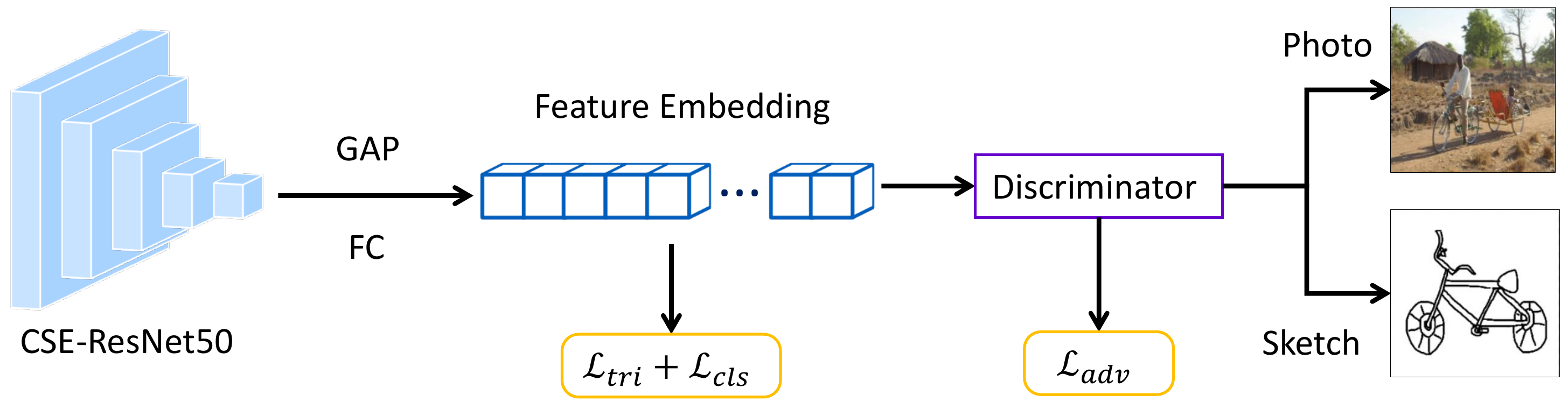}
	\vspace{-0.2cm}
	\caption{
		Network architecture for the generative adversarial network.
	}
	\label{Fig:adv}\vspace{-0.2cm} 
\end{figure}

% \subsubsection{} 
%In this section, 
% We will experimentally validate the superiority of MATHM over another popular feature alignment method
We will experimentally validate the effectiveness of MATHM by comparing it with another popular feature alignment method: the \textbf{Generative Adversarial Network} (GAN). As Figure \ref{Fig:adv} shows, we build up the GAN based on our baseline: we add a fully connected layer after the embedding layer as the discriminator $D$ and use all the other parts of our baseline as the generator $G$. Given an input sample, the generator $G$ tries to conceal the actual modality in the deep feature space, while the discriminator tries to distinguish whether the feature is generated from a sketch or a real photo. Thus the generator tends to produce features that have similar distribution for input sketches and photos. 
The detailed implementation of GAN is described in Appendix \ref{Sec:gan}. %the supplementary material

% \paragraph {Experiment Setup}
Section \ref{sec:MATHM} points out that the model's performance on SBIR task is closely related to two properties of the embedding space: 1) the modality gap and 2) the between-class discrepancy. In this paragraph , we will figure out how generative adversarial training and MATHM affect the model's performance on SBIR. We make a quantitative analysis of the modality gap by calculating the mean difference between the average cosine similarity of same-modality and cross-modality samples of each testing class. And the between-class discrepancy is measured by the mean difference of the average cosine similarity of same-class samples and different-class samples on the testing set.

As results shown in Table \ref{Tab:sim-gap}, the GAN achieves the best performance on reducing the modality gap between the same class samples, but it performs worst on enlarging the between-class discrepancy for both the same-modality and cross-modality samples. For which it gets the lowest mAP and Precision on the SBIR benchmark. Our baseline performed slightly better than others on the between-class discrepancy for same-modality samples, but it has the largest modality gap, which compromises its performance on SBIR. In contrast, the proposed MATHM performs reasonably well on reducing the modality gap as well as enlarging the between-class discrepancy, achieving the best performance on the SBIR benchmarks. This again validates that reducing the modality gap and enlarging the between-class discrepancy is of equal importance for the SBIR task. % in training a good SBIR model. % , missing any one of them will lead to a significant performance drop.

\subsection {Ablation Study}
%\paragraph {Ablation Study.}

%\paragraph {Hyper-parameter Analysis}
%We analyze the effect of hyper-parameter $\lambda$ on the TU-Berlin Extension dataset. Setting $\lambda$ to 0 means the model is trained without the embedding loss. Table \ref{tab:lambda1} shows that mAP@all and Prec@100 scores reach a peak value when $\lambda=1.0$. Thus we set $\lambda$ to 1.0 in all the other experiments.

% In Table \ref{tab:loss2}, we analyze the effect on each of the three kind of triplet relationship.
\label{sec:ablation}
%\paragraph {Model Ablation}
% We conduct extensive ablation studies on different compositions of embedding loss to validate our method, 
% Extensive ablation studies are conducted on each component to validate our method, We conduct ablation studies on each component to validate our method, where the results shown in %, where the results are shown in Table \ref{tab:loss}. 
In this section, we will focus on evaluating the effectiveness of each component of our method. 
%This paragraph evaluates the effectiveness of each component of our method. 
The experimental results are shown in Table \ref{tab:loss}, where the $\mathcal L_{cross}$, $\mathcal L_{in}$, and $\mathcal L_{hyb}$ stand for the cross-modality triplet loss, within-modality triplet loss, and hybrid triplet loss, respectively. $GW$ denotes using a gradient-based weighting strategy to aggregate the used embedding loss. % Otherwise, they are simply summed up together. In all these experiments, a softmax cross-entropy loss is used as the classification loss. 

%\begin{table}[t]
%	\renewcommand\arraystretch{1.1}
%	\centering
%	\small
%	\caption{
%		% hyper-parameter analysis on $\lambda$.
%		The mAP@all and Prec@100 results on the TU-Berlin Extension dataset with different $\lambda$.
%	}\vspace{-0.2cm}
%	\begin{tabular}{lcc}
%		\hline
%		\multicolumn{1}{c}{$\lambda$} & mAP@all        & Prec@100       \\ \hline
%		0.0        & 46.00 $\pm$   0.16        & 56.42 $\pm$   0.18  \\
%		0.5        & 49.98 $\pm$ 0.19 & 60.12 $\pm$ 0.25 \\
%		1.0        & \textbf{50.82 $\pm$ 0.12} & \textbf{60.62 $\pm$ 0.18} \\
%		1.5        & 50.58 $\pm$ 0.33 & 60.50 $\pm$ 0.39  \\
%		2.0        & 50.58 $\pm$ 0.27 & 60.30 $\pm$ 0.35  \\
%		2.5        & 50.36 $\pm$ 0.19 & 60.00 $\pm$ 0.24 \\
%		3.0        & 50.38 $\pm$ 0.17 & 60.10 $\pm$ 0.29  \\ \hline
%	\end{tabular}
%	\vspace{-0.4cm} 
%	\label{tab:lambda1}
%\end{table}

To validate the effectiveness of each components, we first train a model to map the sketches and photos into a common embedding space with only the classification loss. Then we enhance the model with the cross-modality triplet loss, within-modality triplet loss and the hybrid triplet loss separately. The results shows that each type of triplet loss improves the mAP and Prec scores by a different margin. The within-modality triplet loss performs slightly worse than the baseline cross-modality triplet loss, while the hybrid triplet loss performs better than our baseline. This implies that the cross-modality relationship is more important than the within-modality relationship in the ZS-SBIR task. But using the two types of relationships separately could not lead to optimal performance. Next, we combine our baseline with the other two types of triplet loss (the within-modality triplet loss and hybrid triplet loss). The combination of all three triplet losses significantly outperforms our baseline, while combine the baseline with the within-modality triplet loss and hybrid triplet loss separately only brings marginal improvement. This experiment confirms that the within-modality triplet loss and hybrid triplet loss complement each other. Missing either one of them will greatly deteriorate the performance. Finally, we introduce the gradient-based weighting scheme to our model for aggregating all these losses, which further improves the retrieval accuracy. 
The above experiments validate that each of the three triplet losses, as well as the weighting strategy, are indispensable for training a strong ZS-SBIR model.  

The analysis of hyper-parameter $\lambda$ and qualitative analysis of MATHM are included in the supplementary material for the limited space.

\section{Limitations}
%This paper proposes a simple and effective framework to tackle the challenging ZS-SBIR task. 
In MATHM, we implicitly assumed that the sketches and photos from the same category should be visually similar. A shared CNN backbone is capable of capturing their visual similarity. The visualization of retrieval in 
Appendix \ref{Sec:vis} shows that our model can correctly retrieve visually similar images to the query sketch but might produce wrong results for some ambiguous inputs, \emph{e.g.}, the sketch of objects with similar contour but different texture or color. How to deal with the ambiguity of the hand-drawn sketches will be explored in our future work.
%This paper tackle \ref{Sec:vis}

\section{Conclusion}

This paper tackles the ZS-SBIR task and makes two contributions. First, we integrate two popular deep metric learning manners, \emph{i.e.}, the classification learning and pairwise learning, to learn a discriminative metric space. It achieves competitive retrieval accuracy without bells and whistles, providing a simple and strong baseline for ZS-SBIR. Second, based on this baseline, we further suppress the modality gap of sketch query and photo gallery using a novel Modality-Aware Triplet Hard Mining (MATHM) method. MATHM enhances the baseline (particularly the pairwise training part) with several triplet forms, namely within-modality triplet and hybrid triplet. By dynamically balancing the contribution of these triplets during training, MATHM effectively suppresses the modality gap and facilitates better 
category discrimination. 
Extensive experiments conducted on TU-Berlin and Sketchy validate the effectiveness of MATHM. The achieved results are on par with the state-of-the-art.

%%%%%%%%% REFERENCES
{\small
\bibliographystyle{ieee_fullname}
\bibliography{egbib}
}

\newpage
\onecolumn
\newpage
%\twocolumn

\begin{appendices}
	
%------------------------------------------------------------------------
\section{} % {Appendix}

\label{Sec:sup}
%-------------------------------------------------------------------------
\subsection{Implementation details of GAN}
\label{Sec:gan}
%% \subsubsection{} 
%In this section, we will experimentally validate the superiority of MATHM over another popular feature alignment method: the \textbf{Generative Adversarial Network}. 
%We build up the generative adversarial network based on our baseline: we add a fully connected layer after the embedding layer as the discriminator $D$ and use all the other parts of our baseline as the generator $G$. Given an input sample, the generator $G$ tries to conceal the actual modality in the deep feature space, while the discriminator tries to distinguish whether the feature is generated from a sketch or a real photo. 
The objective of generator $G$ and discriminator $D$ is written as:
\vspace{-0.0cm}
\begin{equation}
\begin{aligned}
\mathcal L_{adv-D}(G, D, S, P)&=\mathbb{E}_{p\sim pdata(p)}[\log{(D(G(p)))}] \\
&+\mathbb{E}_{s\sim pdata(s)}[\log{(1-D(G(s)))}],
\label{equ:L_adv-d}
\end{aligned}
\end{equation}
\vspace{-0.0cm}
\begin{equation}
\begin{aligned}
\mathcal L_{adv-G}(G, D, S, P)&=\mathbb{E}_{p\sim pdata(p)}[\log{(1-D(G(p)))}] \\
&+\mathbb{E}_{s\sim pdata(s)}[\log{(D(G(s)))}],
\label{equ:L_adv-e}
\end{aligned}
\end{equation}
where $S$ is the input sketches and $P$ is the input photos.
The $\mathcal L_{adv-D}$ teaches the discriminator $D$ to make the right predictions on the modality of input feature,
while the $\mathcal L_{adv-G}$ encourages $D$ to produce the wrong prediction by updating generator $G$. 
% attempt to fool the discriminator $D$ with generator $G$ by lead $D$ to produce the wrong prediction.
During each iteration, we fix $D$ and update $G$ with $\mathcal L_{all}$ (Eq. \ref{equ:all}) and $\mathcal L_{adv-G}$ at first, then freeze $G$ and update $D$ with $\mathcal L_{adv-D}$.
By optimizing $G$ and $D$ with $\mathcal L_{adv-G}$ and $\mathcal L_{adv-D}$ iteratively,
the generator tends to produce features that have similar distribution for input sketches and photos. % Thus the modality gap is minimized.

\subsection {Hyper-parameter Analysis}
We analyze the effect of hyper-parameter $\lambda$ on the TU-Berlin Extension dataset. Setting $\lambda$ to 0 means the model is trained without the embedding loss. Table \ref{tab:lambda1} shows that mAP@all and Prec@100 scores reach a peak value when $\lambda=1.0$. Thus we set $\lambda$ to 1.0 in all the other experiments.

\begin{table}[h]
	\renewcommand\arraystretch{1.1}
	\centering
	\small
	\caption{
		% hyper-parameter analysis on $\lambda$.
		The mAP@all and Prec@100 results on the TU-Berlin Extension dataset with different $\lambda$.
	}\vspace{-0.2cm}
	\begin{tabular}{lcc}
		\hline
		\multicolumn{1}{c}{$\lambda$} & mAP@all        & Prec@100       \\ \hline
		0.0        & 46.00 $\pm$   0.16        & 56.42 $\pm$   0.18  \\
		0.5        & 49.98 $\pm$ 0.19 & 60.12 $\pm$ 0.25 \\
		1.0        & \textbf{50.82 $\pm$ 0.12} & \textbf{60.62 $\pm$ 0.18} \\
		1.5        & 50.58 $\pm$ 0.33 & 60.50 $\pm$ 0.39  \\
		2.0        & 50.58 $\pm$ 0.27 & 60.30 $\pm$ 0.35  \\
		2.5        & 50.36 $\pm$ 0.19 & 60.00 $\pm$ 0.24 \\
		3.0        & 50.38 $\pm$ 0.17 & 60.10 $\pm$ 0.29  \\ \hline
	\end{tabular}
	\vspace{-0.4cm} 
	\label{tab:lambda1}
\end{table}

%-------------------------------------------------------------------------
\subsection{Qualitative Analysis}

\label{Sec:vis}

\begin{figure*}[h]
	\centering
	\quad
	
	\begin{subfigure}{0.44\linewidth}
		\includegraphics[width=\linewidth]{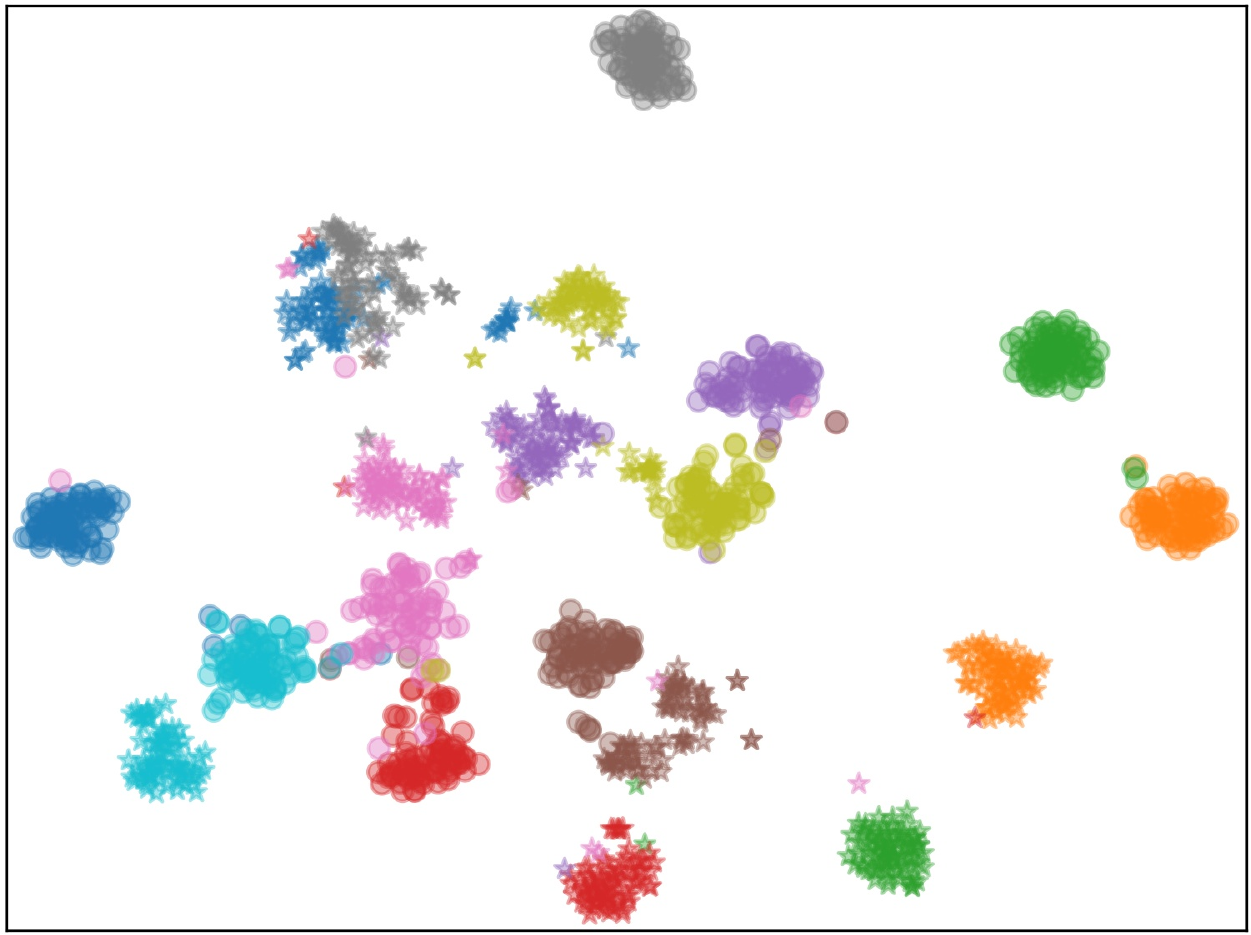}
		\caption{Baseline}
		\label{fig:base}
	\end{subfigure}
	\quad
	\quad
	\begin{subfigure}{0.44\linewidth}
		\includegraphics[width=\linewidth]{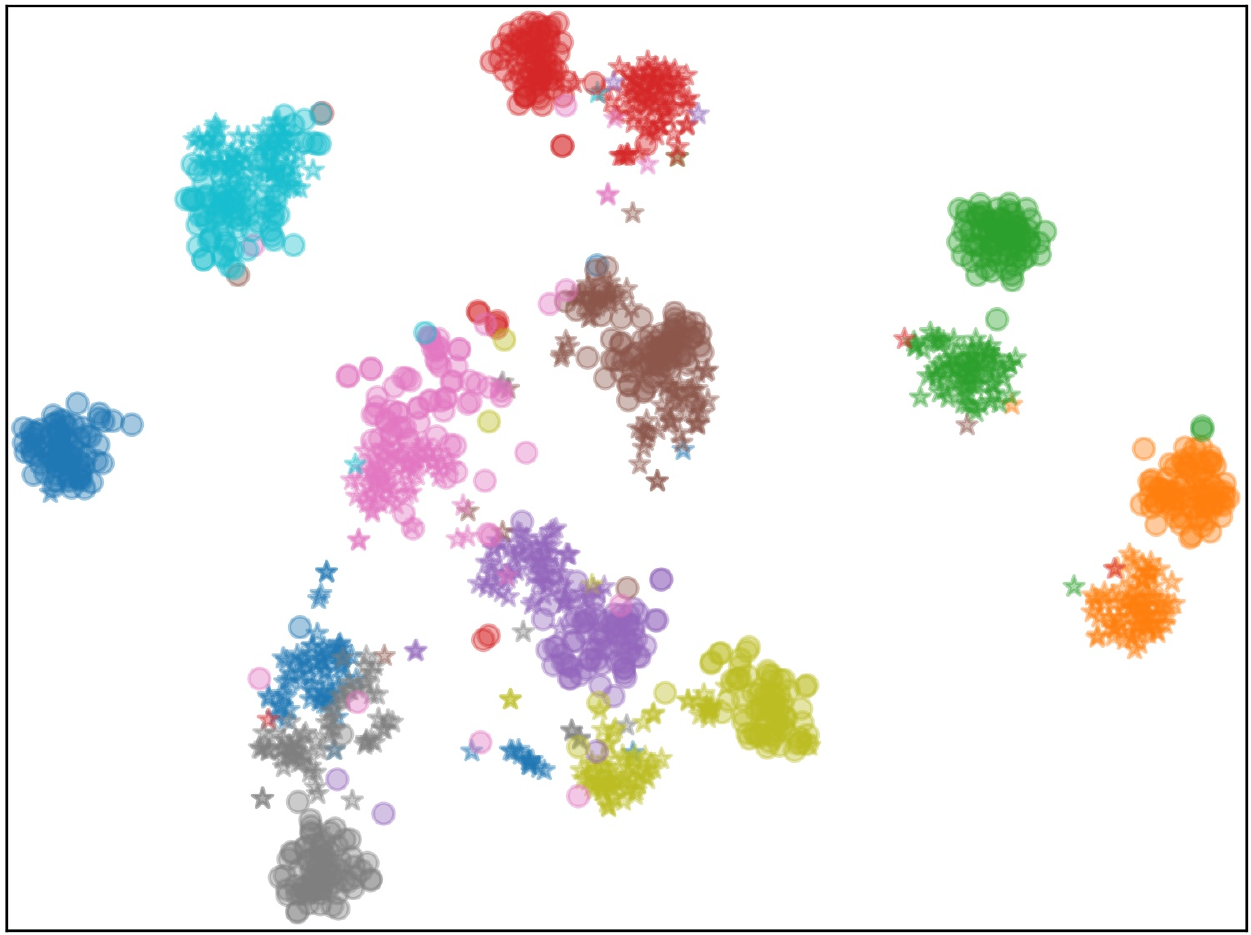}
		\caption{MATHM}
		\label{fig:all}
	\end{subfigure}
	%	\subfigure[Baseline]{
	%		\includegraphics[width=0.44\linewidth]{Figures//tsne-base.pdf}
	%	}\vspace{-0.2cm}
	%	\quad
	%	\quad
	%	\subfigure[MATHM]{
	%		\includegraphics[width=0.44\linewidth]{Figures//tsne-all.pdf}
	%	}
	\caption{The t-SNE \cite{van2008visualizing} visualization of the feature representation on ten randomly chosen classes from the testing set of TU-Berlin Extension. (a) and (b) show the features extracted by our baseline and MATHM, respectively. Each color represents a particular class. The stars denote sketches, and circles denote photos. Best viewed in color.
	}
	\label{Fig:vis}\vspace{-0.4cm} 
\end{figure*}

\paragraph {Visualization of Feature Embedding}
We visualize the 512-d feature extracted by our baseline model and the full model on the test set of TU-Berlin Extension in Figure \ref{Fig:vis}. 
The feature representation of our MATHM is well clustered by categories. In contrast, our baseline model results show a large modality gap between samples with different modalities of the same category. 
% We also performed quantitative analysis on this issue by calculating the average L2 distance of cross-modality and same-modality samples on TU-Berlin. As results shown in Tab \ref{Tab:dist}, the within-class distance of cross- (same-)modality is $1.211$ ($0.995$) and $1.150$ ($1.000$) for baseline and MATHM, respectively. It indicates that MATHM reduces the modality gap from $0.216$ to $0.150$, and is consistent with our qualitative observation in Fig. \ref{Fig:vis}.
The visualization is consistent with our qualitative analysis, which validates that MATHM resolves the problem of modality discrepancy better than our baseline.

\begin{figure}[t]
	\centering
	\includegraphics[width=0.56\linewidth]{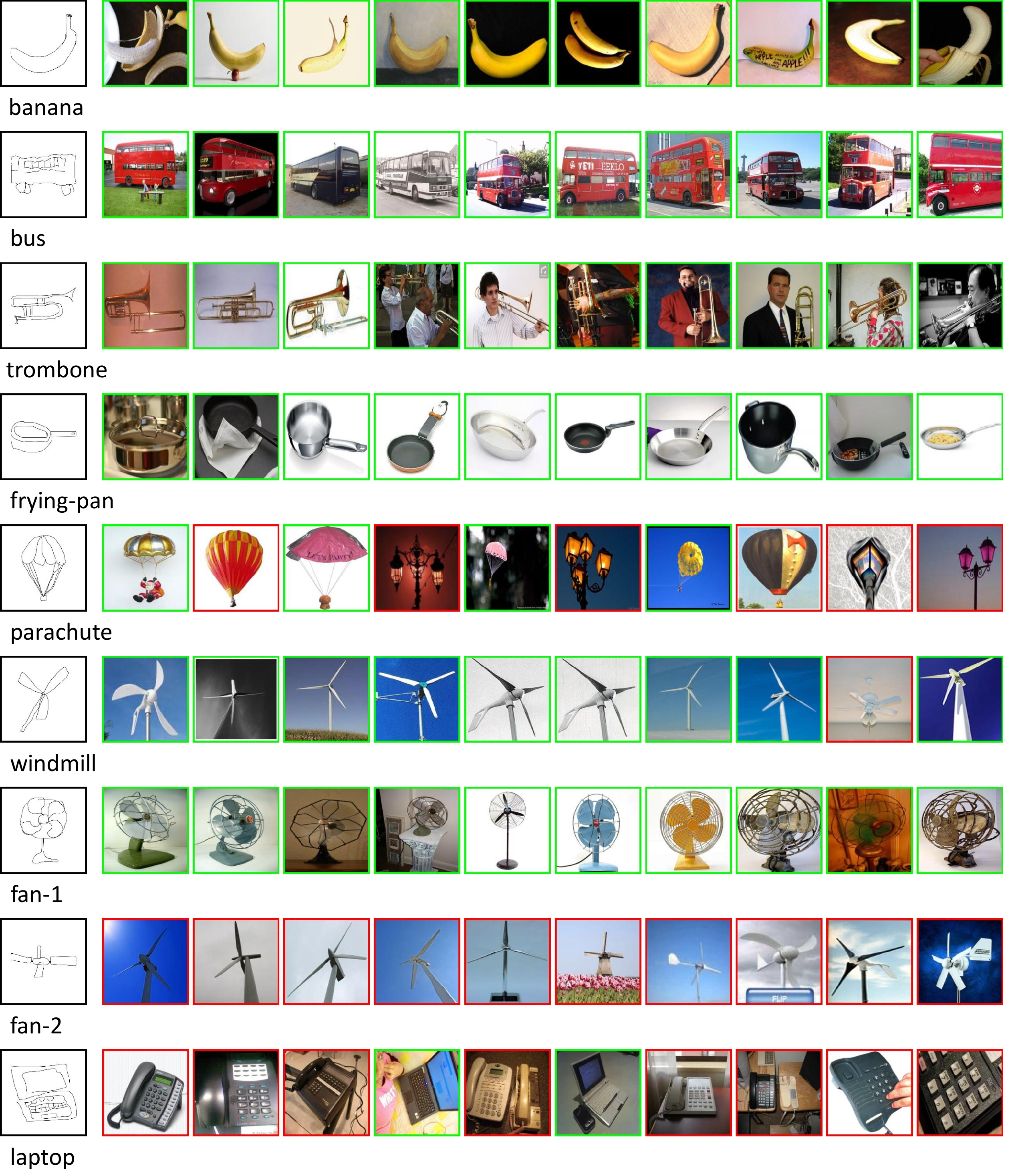}
	
	\caption{Top-10 retrieval results of our full model on TU-Berlin Extension. The correct results are in green, while the wrong results are in red. Best viewed in color.}
	\label{Fig:rank}\vspace{-0.4cm} 
\end{figure}

\paragraph {Visualization of Retrievals}
Figure \ref{Fig:rank} shows the Top-10 retrieval results obtain by our full model on the TU-Berlin Extension dataset. Our model successfully retrieves photo images with the correct category label in most cases. 
We also present some negative cases where our model fails to retrieve photos with the same category as the query sketch. These wrongly retrieved images usually share some salient visual component to the ground truth categories, \emph{e.g.}, both the \emph{fan} and \emph{windmill} contain three pieces of blades while the \emph{laptop} and \emph{telephone} have a similar keyboard. If we add a circle on the boundary of the sketch \emph{fan}, as ``fan-1'' in Figure \ref{Fig:rank} shows, the retrieval accuracy dramatically increases. This important observation implies that our model has learned a shared embedding space for inputs from both sketches and photos. The failure of retrieval is mainly due to the inter-class similarity of sketches from certain categories.

%\end{document}

\end{appendices}

\end{document}